\documentclass[conference, twocolumn]{IEEEtran}
\IEEEoverridecommandlockouts

\usepackage[
letterpaper,      
top=1in,          
bottom=0.75in,     
left=0.75in,       
right=0.75in,     
columnsep=0.25in   
]{geometry}

\usepackage{cite}
\usepackage{amsmath,amssymb,amsfonts}
\usepackage{algorithmic}
\usepackage{graphicx}
\usepackage{textcomp}
\usepackage{xcolor}
\usepackage{colortbl}
\usepackage{multirow}
\usepackage{booktabs}
\usepackage{url}
\usepackage[ruled,vlined]{algorithm2e}
\graphicspath{{./Figures/}}

\title{\LARGE \bf LLM-Enhanced Spatio-temporal Learning for Grid-Level Docked Bike Sharing Demand Prediction}

\author{
	\parbox{\textwidth}{%
		\centering
		Xuxilu Zhang$^{1}$ and Francesc Soriguera$^{1}$
	}%
	\thanks{$^{1}$ BIT -- Universitat Polit\`ecnica de Catalunya --
		Barcelona Tech (UPC), Jordi Girona 1--3, Building B1, Office 111,
		Barcelona 08034, Spain.
		{\tt\small \{xuxilu.zhang, francesc.soriguera\}@upc.edu}}
}
\begin{document}
	
	\maketitle
	\thispagestyle{empty}
	\pagestyle{empty}
	
	\begin{abstract}
		Short-term bike-sharing demand forecasting is complicated by spatial-temporal non-stationarity and the practical difficulty of incorporating unstructured external text into numerical pipelines. Conventional approaches rely on historical flow sequences and fixed graph structures, thereby constraining their accuracy when anomalous social events perturb normal travel patterns.  We propose a forecasting framework in which a Large Language Model (LLM) drives a semantic shockwave mechanism that converts free-form urban text—municipal event schedules, local news, and transit bulletins—into quantified spatial-temporal perturbation fields. The LLM extracts three physically interpretable parameters per event (intensity, spatial reach, and temporal lag), from which Gaussian decay fields are constructed and injected into a Zero-Inflated Adaptive Spatio-Temporal Graph Convolutional Network (ZI-ASTGCN). To handle the pronounced sparsity of grid-level measurements, the model couples a dual-branch output head with a multi-task zero-inflated loss that jointly trains a gating probability and a conditional flow intensity. Experiments on the operational Barcelona Bicing dataset show that ZI-ASTGCN outperforms established neural baselines, with particularly strong gains during high-demand periods, validating the utility of physics-grounded semantic signals in spatial-temporal mobility forecasting.
	\end{abstract}
	
	\section{Introduction}
	\label{sec:introduction}
	
	Reliable short-term demand forecasting is a prerequisite for fleet rebalancing, station capacity management, and network robustness in Bike-Sharing Systems (BSS)~\cite{soriguera2020continuous,jimenez2024continuous,zhang2025evaluating}. As micro-mobility networks expand in dense urban areas, forecasting accuracy directly conditions operational efficiency at the spatial resolution of individual docking stations. Yet fine-grained demand prediction faces two persistent obstacles: spatial-temporal non-stationarity and the prevalence of near-zero observations in low-activity grid cells~\cite{subramanian2023enhancing}. Although regular commuting rhythms impose a coarse temporal skeleton on aggregate demand, abrupt shocks driven by large public events, transit disruptions, or sudden weather changes routinely break this regularity~\cite{yang2020using,torres2024forecasting,patil2025context}. During such intervals, grid-level flow records become zero-inflated and heavy-tailed, straining both model fit and rebalancing decisions~\cite{jimenez2024optimization,jiang2025data}.
	
	State-of-the-art spatial-temporal models—including graph convolutional and attention-based architectures—are fundamentally closed numerical systems~\cite{moghadas2024long,li2024spatio}. They aggregate historical flow tensors and exploit static spatial topologies, but carry no mechanism to interpret free-form contextual narratives. The consequence is that demand surges driven by large-scale events appear indistinguishable from random noise until their effects are already visible in the observation stream~\cite{liu2024spatial,wang2025gsf}. By the time the model detects an anomaly, the operational window for anticipatory rebalancing has often closed.
	
	Recent work on Large Language Models (LLMs) has demonstrated their capacity to distill unstructured text into features compatible with time-series models~\cite{zhang2024large,wen2025foundation}. Several studies tokenize multi-modal traffic sequences for transformer-based encoding~\cite{liu2024spatial,jin2024time}, while others align text-derived embeddings directly with sensor observations~\cite{wang2025gsf,liu2025timecma}. A common limitation of these approaches is that they encode semantic information as static latent vectors, without modeling how an event's influence spreads outward from its origin and attenuates over distance and time. For instance, the influence of a stadium concert does not instantaneously materialize across the city; it propagates gradually and decays with spatial separation and temporal distance from the event peak.
	
	We address this gap by proposing a framework that models urban events as parameterized physical impulses. A reasoning-focused LLM parses each narrative text entry into three scalar parameters—event intensity, spatial impact radius, and temporal lag—which are then used to construct continuous Gaussian decay fields on an hourly spatial grid. These semantic shockwave fields serve as conditioning signals for a Zero-Inflated Adaptive Spatio-Temporal Graph Convolutional Network (ZI-ASTGCN), which extends the baseline ASTGCN~\cite{guo2019attention} with a cross-attention fusion layer and a dual-branch prediction head. The dual-branch design simultaneously models the probability of demand occurrence (via a learned gating mechanism) and the conditional demand magnitude (via a Huber-loss regressor), directly addressing the zero-inflated and heavy-tailed nature of fine-grained mobility observations.
	
	The main contributions of this study are:
	\begin{itemize}
		\item We propose a semantic shockwave framework that derives explicit physical parameters from unstructured urban text using an LLM, transforming narrative event descriptions into continuous spatio-temporal Gaussian decay fields that integrate seamlessly with graph convolutional architectures.
		\item We introduce ZI-ASTGCN, which incorporates these semantic fields using a cross-attention fusion module and a dual-branch output layer trained with a multi-task zero-inflated objective, enabling a clear disentanglement between demand occurrence and demand intensity in sparse, highly skewed data.
		\item We conduct an extensive empirical study on the real-world Barcelona Bicing dataset, including benchmarks against spatio-temporal neural baselines, an internal ablation focused on the LLM component, and a dedicated event-period error metric (EvtRMSE) designed to isolate model accuracy during active semantic shockwave intervals.
	\end{itemize}
	
	\section{Methodology}
	\label{sec:methodology}
	
\begin{figure}[t!]
		\centering
		\includegraphics[width=\columnwidth]{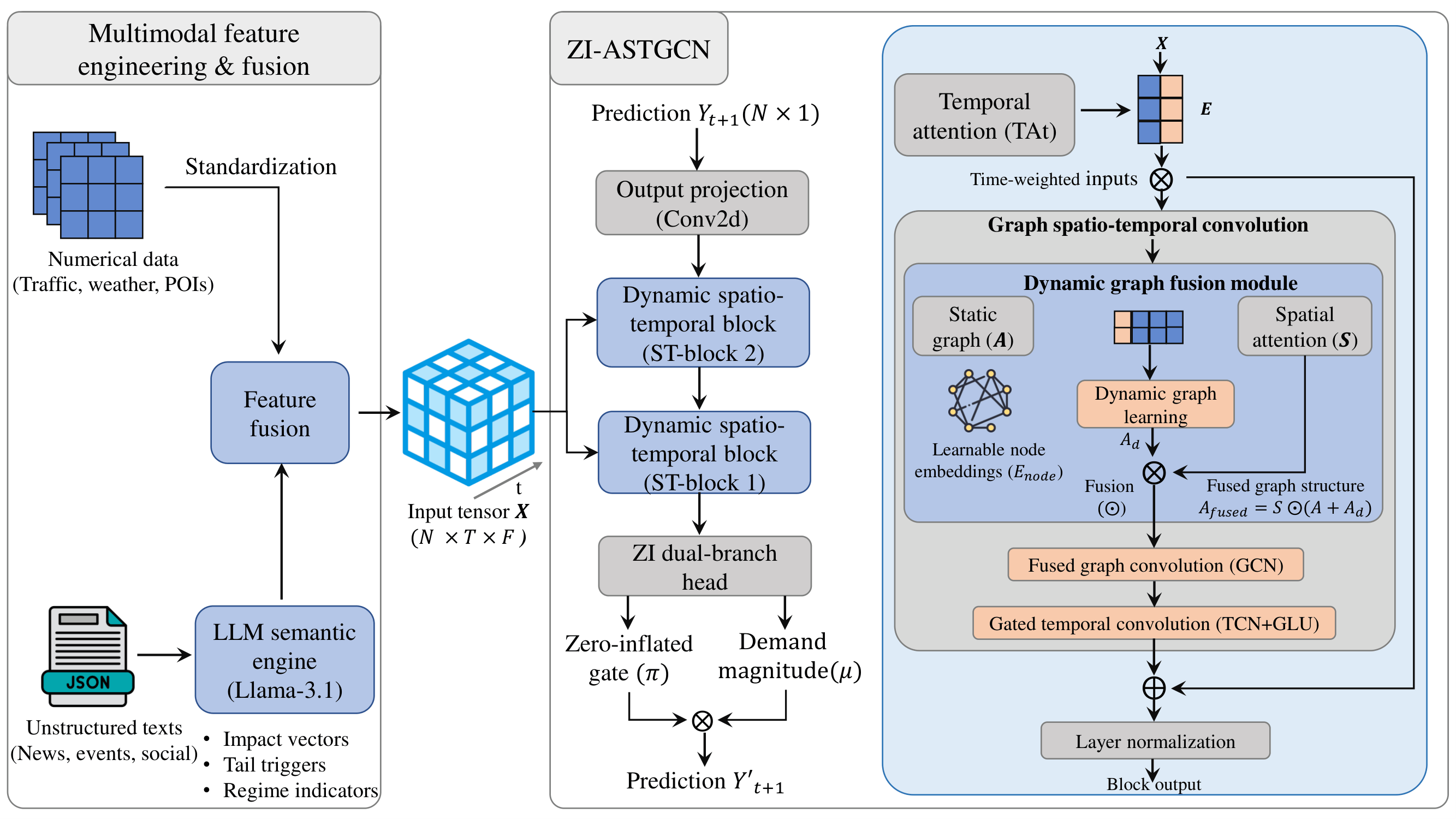}
		\caption{System architecture. \emph{Left}: multi-modal feature construction. \emph{Center}: ZI-ASTGCN backbone. \emph{Right}: graph convolution processing detail.}
		\label{fig:framework}
	\end{figure}

	The proposed architecture rests on the hypothesis that urban mobility is driven by two main components: a regular periodic component for habitual commuting and a transient component linked to specific external events. Figure~\ref{fig:framework} shows the pipeline. An LLM semantic engine transforms narrative text into structured parameters, which then adjust the attention layers and feature space of the ZI-ASTGCN backbone.
	
	\subsection{Problem Formulation}
	
	The bike-sharing network is modelled as a dynamic graph signal over a weighted topology $\mathcal{G} = (\mathcal{V}, \mathcal{E}, \mathbf{A})$, where $\mathcal{V} = \{v_1, \ldots, v_N\}$ is a set of $N$ spatial grid cells and $\mathbf{A} \in \mathbb{R}^{N \times N}$ encodes static geographic adjacency.  The network state at time $t$ is a feature tensor $\mathbf{X}_t \in \mathbb{R}^{N \times C}$, constructed as:
	\begin{equation}
		\mathbf{X}_t = \mathrm{Concat}\!\left(
		\mathbf{X}^{\mathrm{flow}}_t,\; \mathbf{X}^{\mathrm{sem}}_t
		\right),
	\end{equation}
	
	where $\mathbf{X}^{\mathrm{flow}}_t$ holds inflow and outflow volumes together with hourly weather covariates and $\mathbf{X}^{\mathrm{sem}}_t$ carries the semantic impact field produced by the shockwave module (Section~\ref{subsec:shockwave}).
	
	Because many grid cells record zero flows during off-peak hours, the target prediction $\hat{y}_{i,t+1}$ is factored as:
	\begin{equation}
		\hat{y}_{i,t+1} = g_{i,t+1} \cdot v_{i,t+1},
	\end{equation}
	
	where $g_{i,t+1} \in (0,1)$ is a gating probability (demand occurrence) and $v_{i,t+1} \geq 0$ is a conditional intensity (demand volume given occurrence).  Both are produced by parallel output branches sharing a common latent representation.  Given a look-back window of $T_{\mathrm{in}}$ hours, the model learns a mapping $\mathcal{F}_\Theta : \mathbb{R}^{N \times T_{\mathrm{in}} \times C} \to \mathbb{R}^{N \times 1}$ by minimising the multi-task zero-inflated loss described in Section~\ref{subsec:loss}.
	
	\subsection{Physics-Informed Semantic Shockwave}
	\label{subsec:shockwave}
	
	Standard embedding models, like BERT and sentence transformers, focus on lexical proximity but lack guidance on how urban disturbances spread spatially and decay over time. The semantic shockwave mechanism improves upon this by treating each textual event as a discrete impulse that spreads through a spatial grid using a Gaussian decay function.
	
	\subsubsection{Physical Parameter Extraction}
\begin{figure}[t!]
		\centering
		\includegraphics[width=\columnwidth]{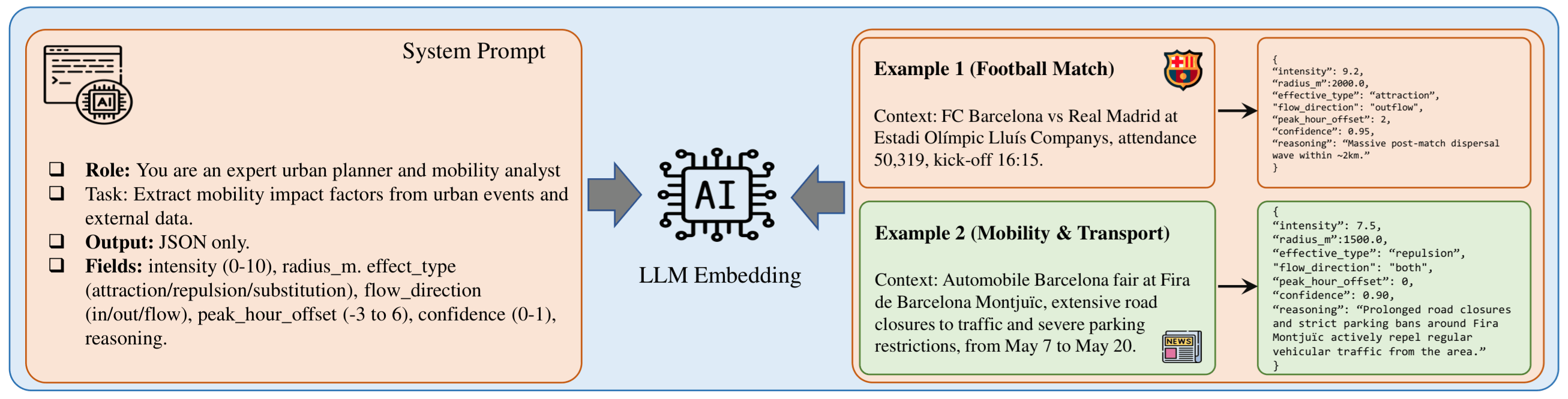}
		\caption{Execution flow of the LLM semantic parsing module.}
		\label{fig:llm_embedding}
	\end{figure}

	As illustrated in Fig.~\ref{fig:llm_embedding}, the LLM is conditioned with a domain-specific system prompt and a set of few-shot exemplars that enforce a structured JSON output schema. For each detected event record $e$, the model returns a parameter tuple $\mathcal{P}_e = \{S_e,\, R_e,\, \Delta t_e\}$: the dimensionless impact intensity score $S_e \in [0,1]$, the effective influence radius $R_e$ (meters), and the temporal offset $\Delta t_e$ relative to the current prediction window. The JSON schema enforces type constraints and value ranges, and outputs failing validation are redirected to a keyword-based fallback that assigns conservative default parameters. This two-tier extraction pipeline preserves downstream robustness without requiring fine-tuning of the base LLM.
	
	\subsubsection{Spatio-Temporal Propagation Field}
	The semantic perturbation induced by event $e$ on cell $i$ at time step $t$ is modeled as:
	\begin{equation}
		I_{i,t}^{(e)} = S_e
		\cdot \exp\!\left(-\frac{d_{i,e}^2}{R_e^2}\right)
		\cdot \exp\!\left(-\frac{\Delta t_e}{\tau}\right),
		\label{eq:shockwave}
	\end{equation}
	where $d_{i,e}$ is the Euclidean distance from the cell centroid to the event epicenter and $\tau$ is a temporal decay constant set to 2.0 hours on the validation set.  Aggregating Eq.~\eqref{eq:shockwave} over all active events yields the composite semantic field $\mathbf{X}^{\mathrm{sem}}_t$, which enters a dedicated feature branch that bypasses global graph convolutions and injects adjustment terms directly into the dual output heads.  This branch is zero-initialized, ensuring exact equivalence to the non-LLM baseline when no events are present.
	
	\begin{figure}[t!]
		\centering
		\includegraphics[width=\columnwidth]{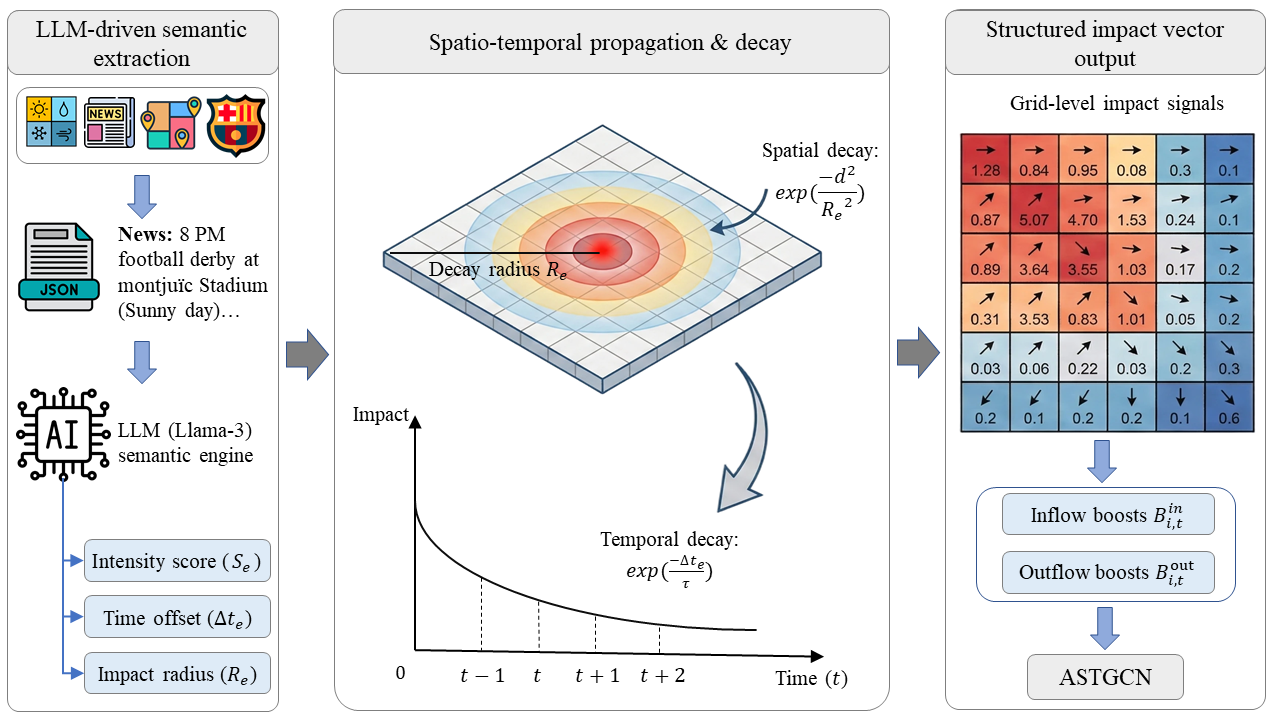}
		\caption{Schematic of the semantic shockwave mechanism.}
		\label{fig:shockwave}
	\end{figure}
	
	\subsection{Dynamic Spatio-Temporal Graph Learning}
	
	Spatial dependencies are resolved over a fused topology:
	\begin{equation}
		\mathbf{A}_{\mathrm{fused}} =
		\mathbf{S} \odot \bigl(\mathbf{A} + \mathbf{A}_d\bigr),
	\end{equation}
	where $\mathbf{A}$ denotes the static geographical adjacency, $\mathbf{A}_d = \mathrm{Softmax}(\mathrm{ReLU}(\mathbf{E}_{\mathrm{node}} \mathbf{E}_{\mathrm{node}}^\top))$ represents a learnable latent graph derived from node embeddings $\mathbf{E}_{\mathrm{node}} \in \mathbb{R}^{N \times D}$ with Top-$K$ sparsification applied, and $\mathbf{S}$ is a data-adaptive spatial attention matrix inferred from the current input state. During active event intervals, a cross-attention layer leverages the semantic impact fields as query keys to redirect the flow of spatial features, strengthening connections toward nodes whose functional roles are consistent with the ongoing event category.
	
	Spectral graph convolutions are implemented using a Chebyshev polynomial approximation, ensuring linear computational complexity with respect to the number of nodes~\cite{guo2019attention}. Temporal evolution is modeled via a Gated Linear Unit (GLU)–based temporal convolution, which selectively attenuates routine periodic patterns while emphasizing event-related perturbation signals propagated through the semantic branch.
	
	\subsection{Optimization Objective}
	\label{subsec:loss}
	Bike-sharing demand at fine spatial resolution is both zero-inflated and heavy-tailed. We address this jointly through a multi-task, zero-inflated loss:
	\begin{equation}
		\mathcal{L}_{\mathrm{ZI}} =
		\frac{1}{|\Omega|}\sum_{i \in \Omega}
		\ell_{\mathrm{gate}}\!\bigl(g_i,\,\mathbb{I}(y_i > 0)\bigr)
		\;+\;
		\frac{\lambda}{|\Omega_{\mathrm{pos}}|}
		\sum_{i \in \Omega_{\mathrm{pos}}}
		\ell_{\mathrm{reg}}\!\bigl(v_i,\, y_i\bigr),
		\label{eq:loss}
	\end{equation}
	where $\Omega$ represents the full sample set, $\Omega_{\mathrm{pos}} \subseteq \Omega$ containing observations where ground-truth demand ($y_i > 0$). The parameter $\lambda$ balances tasks, while the gating term $\ell_{\mathrm{gate}}$ uses binary cross-entropy loss to differentiate active from inactive cells. The regression term $\ell_{\mathrm{reg}}$ applies adaptive Huber loss to positive observations, penalizing under-prediction during demand surges with an asymmetric weight $\alpha_{\mathrm{tail}} = 3.0$. Training in a normalized observation space mitigates numerical instabilities from log-domain transformations on near-zero counts.
	
	\section{Experiments}
	\label{sec:experiments}
	
	\subsection{Data Acquisition and Description}
	\label{subsec:datasets}
	
	The evaluation dataset covers the period from April 1 to June 30, 2025, in Barcelona, and includes normal commuting patterns as well as major urban events, thus providing a challenging testbed for anomaly-driven demand prediction.
	
	\subsubsection{Bike-sharing system flow}
	Operational records are drawn from 510 active Bicing docking stations at 5-minute resolution.  Rather than reconstructing origin-destination trip trajectories—which are incomplete and license-restricted—we infer demand from signed temporal differences in dock occupancy.  Positive differentials correspond to departures (outflow) and negative differentials to arrivals (inflow).
	
	\subsubsection{Geospatial context and points of interest}
	Static spatial characteristics are extracted from OpenStreetMap (OSM): approximately 79,000 POIs classified into 14 functional categories, including transport nodes, retail hubs, and educational facilities. These features provide a stable representation of land-use heterogeneity that informs the latent graph $\mathbf{A}_d$.
	
	\subsubsection{Meteorological conditions}
	Hourly weather records—surface temperature ($^\circ$C), relative humidity (\%), total precipitation (mm), and mean wind speed (m/s)—are collected from local meteorological stations and time-aligned to the demand tensors.
	
	\subsubsection{Unstructured urban corpora and event signals}
	\label{subsubsec:events}
	The semantic input corpus aggregates municipal event calendars, digital news feeds, and transit management bulletins, cataloguing a range of disruptions from scheduled mass gatherings (football matches, music festivals) to non-recurrent anomalies (transit strikes, road closures).
	
	\subsection{Data Processing}
	\label{subsec:dataprocessing}
	
\begin{figure}[htbp]
		\centering
		\includegraphics[width=\columnwidth]{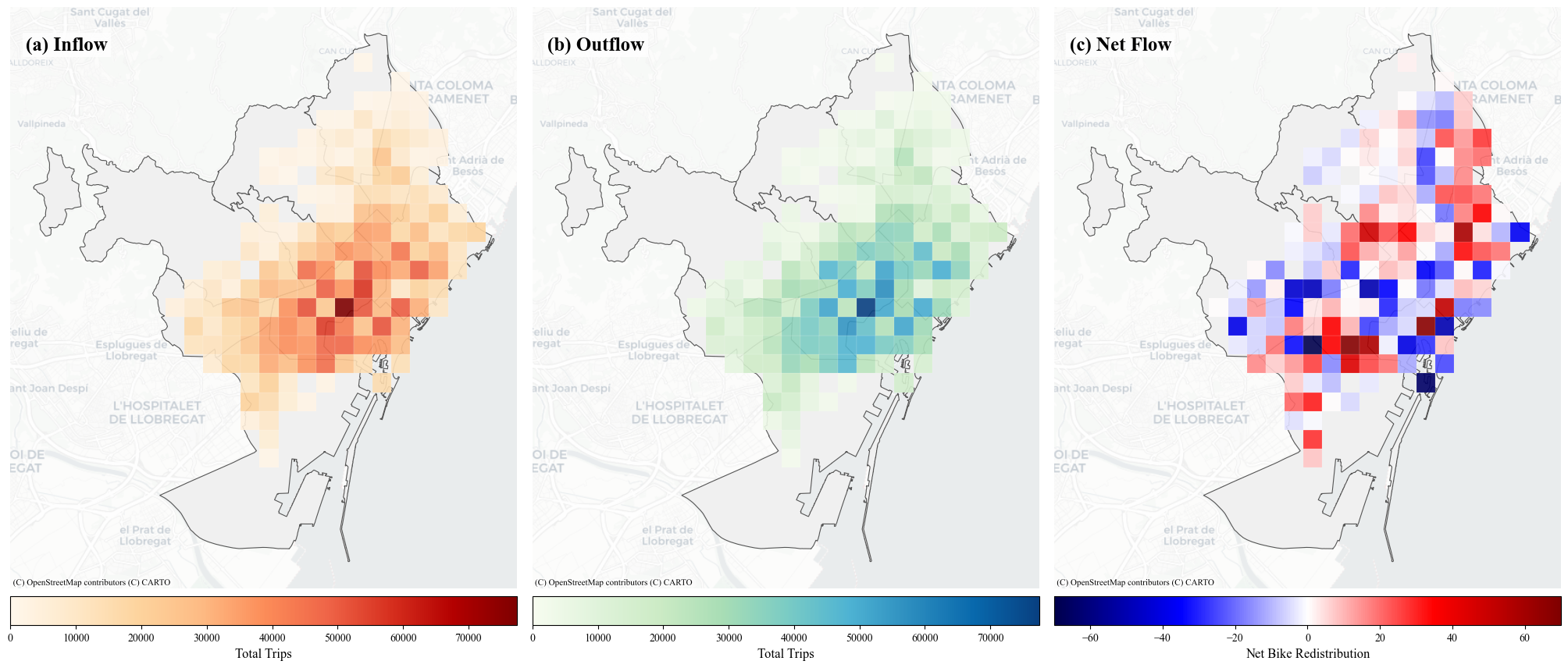}
		\caption{Spatial heterogeneity of accumulated traffic flow across the Barcelona metropolitan grid.  High-demand zones (dark red) cluster around transit terminals and commercial centers.}
		\label{fig:flow_distribution}
	\end{figure}

	The study area is tessellated into $N = 374$ valid grid cells of $800\,\text{m} \times 800\,\text{m}$ each.  Five-minute station observations are mapped to their enclosing cells and aggregated into 1-hour bins to form the inflow and outflow tensors.  The maximum aggregate dock capacity $C_i$ within each cell is computed from historical records and applied as a hard physical upper bound on predicted flows.  Figure~\ref{fig:flow_distribution} illustrates the resulting spatial heterogeneity: demand concentrates heavily around central transit hubs and commercial corridors, with peripheral residential cells recording frequent zero intervals.
	
	\subsection{Semantic Feature Engineering}
	\label{subsec:feature_engineering}
	
\begin{figure}[htbp]
		\centering
		\includegraphics[width=\columnwidth]{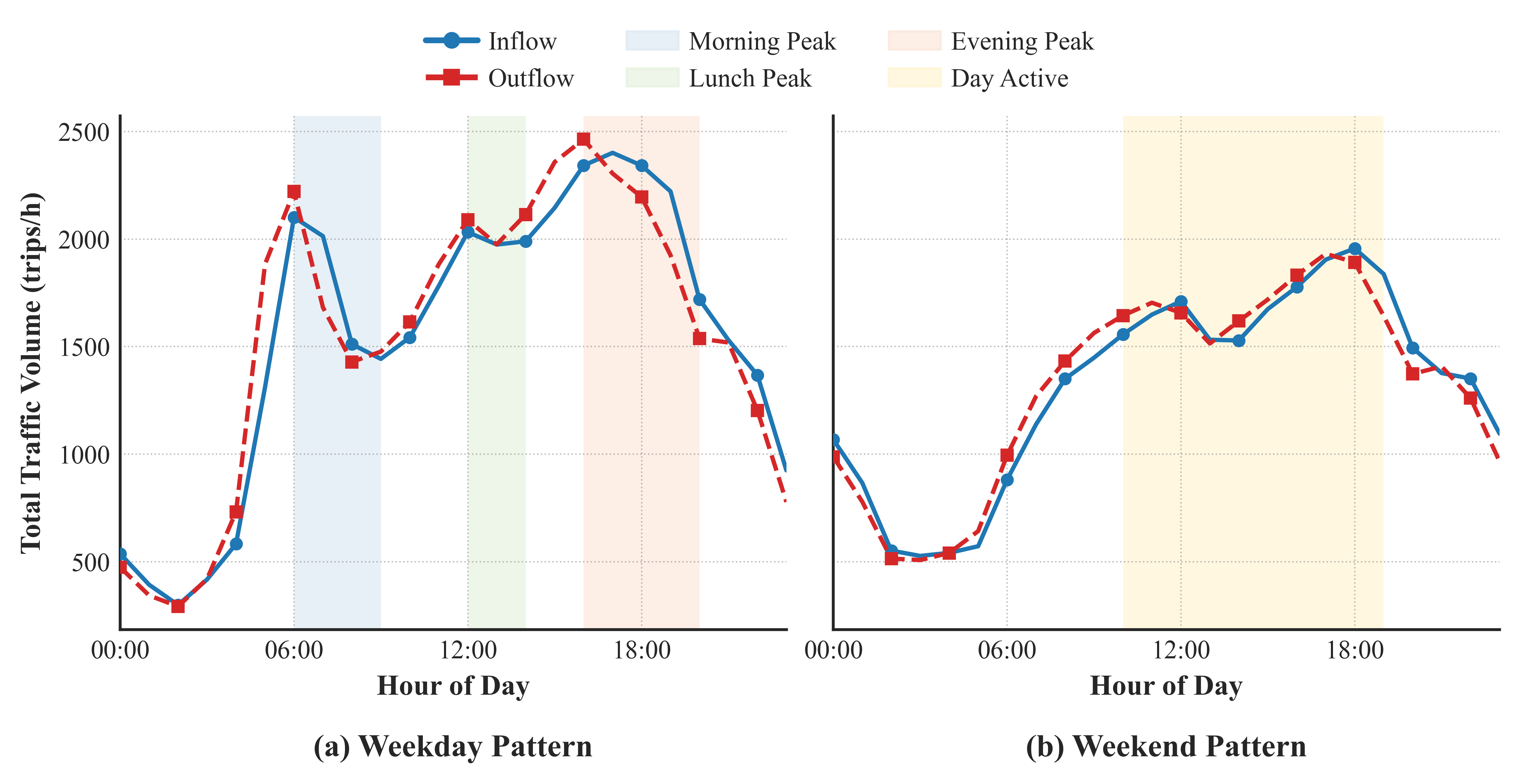}
		\caption{Aggregate bike-sharing demand by hour for weekdays (left) and weekends (right), revealing distinct morning, lunch, and evening peaks on weekdays and a broader midday plateau at weekends.}
		\label{fig:hourly_pattern}
	\end{figure}

	Each narrative event record is processed through the LLM extraction pipeline from Section~\ref{subsec:shockwave} to obtain the parameter tuple $(S_e, R_e, \Delta t_e)$. The resulting decay fields $\mathbf{X}^{\mathrm{sem}}_t$ are calculated hourly for all $N$ cells and combined with the numerical feature tensor. Additionally, binary peak-hour masks and weekend flags are added to reflect the daily and weekly structure shown in Fig.~\ref{fig:hourly_pattern}.
	
	\subsection{Implementation and Training Protocol}
	\label{subsec:implementation}
	
	The ZI-ASTGCN model processes 12 hours of historical data ($T_{\mathrm{in}} = 12$) to predict the demand for the next hour. It features a hidden channel size of $d_{\mathrm{hidden}} = 64$ and retains the Top-$K = 24$ neighbors per node to reduce over-smoothing. With a spatial dropout of 0.2 and a tail penalty coefficient $\alpha_{\mathrm{tail}} = 3.0$, the model is optimized using AdamW with an initial learning rate of $10^{-3}$, applying a loss weight of 2.0 for active samples. Training lasts up to 200 epochs with early stopping after 30 epochs.
	
	The dataset is split chronologically: 70\% for training, 10\% for validation, and 20\% for testing. Performance is evaluated using RMSE, MAE, $R^2$, and WAPE, with additional metrics for robustness—Tail-RMSE for the upper 90th percentile of volumes and Event-RMSE (EvtRMSE) for time steps with active shockwave signals. EvtRMSE primarily compares the full ZI-ASTGCN to its non-LLM variant, highlighting semantic reasoning benefits during disruptions.
	
	\section{Results and Discussion}
	\label{sec:result}
	
	\subsection{Comparative Performance Analysis}
	
	Table~\ref{tab:model_performance} presents results against three standard spatio-temporal neural baselines: STGCN~\cite{yu2018spatio}, DCRNN~\cite{li2018diffusion}, and GraphWaveNet~\cite{wu2019graph}. All baselines are trained with the same feature matrix and data split as ZI-ASTGCN to ensure a fair, controlled comparison. We also include an ablated configuration (ZI-ASTGCN w/o LLM) that removes the semantic shockwave fields to isolate the effect of the LLM component.
	
	For the inflow task, ZI-ASTGCN with LLM embeddings achieves the lowest error across all metrics, with an RMSE of 3.112 compared with DCRNN’s 3.254 and a 31\% reduction in MAE (1.360 vs.\@ 1.973), indicating that the model improves typical predictions across the distribution rather than only mitigating a few large errors. This pattern stems from the dual-branch architecture: conventional neural baselines optimize a single continuous loss over all samples, which encourages small non-zero outputs even for inactive grid cells and consequently inflates MAE in sparse regions. In contrast, ZI-ASTGCN’s gating branch explicitly suppresses low-activity cells, allowing the regression branch to focus on periods with meaningful demand.
	
	The incremental benefit of the LLM module is most apparent in the event-focused metrics. Compared with the non-LLM variant, the full model lowers inflow EvtRMSE from 3.653 to 3.359 and outflow EvtRMSE from 3.470 to 3.076, demonstrating that the semantic fields introduce anticipatory signals beyond those contained in historical flows alone. Tail-RMSE further declines to 7.631 (inflow) and 7.679 (outflow), aligning with the asymmetric tail loss that steers gradient updates toward under-predicted surges. One caveat is that STGCN attains a slightly better outflow Tail-RMSE (7.469) than the full model (7.679); this isolated exception may arise from STGCN’s spectral filters aligning with a specific frequency band of outflow spikes and merits additional analysis.
	
	\begin{table}[htbp]
		\centering
		\caption{Performance Comparison with State-of-the-Art Baselines}
		\label{tab:model_performance}
		\renewcommand{\arraystretch}{1.2}
		\setlength{\tabcolsep}{2.5pt}
		\resizebox{\columnwidth}{!}{%
			\begin{tabular}{l c c c c c c c}
				\toprule
				\textbf{Model} & \textbf{Target} &
				\textbf{RMSE} & \textbf{MAE} & \textbf{R$^2$} &
				\textbf{WAPE} & \textbf{Tail-RMSE} & \textbf{EvtRMSE} \\
				\midrule
				\multirow{2}{*}{STGCN}
				& Inflow  & 3.319 & 2.082 & 0.841 & 0.493 & 8.348 & -- \\
				& Outflow & 3.136 & 2.016 & 0.857 & 0.477 & \textbf{7.469} & -- \\
				\midrule
				\multirow{2}{*}{DCRNN}
				& Inflow  & 3.254 & 1.973 & 0.847 & 0.467 & 8.227 & -- \\
				& Outflow & 3.234 & 1.955 & 0.848 & 0.462 & 8.160 & -- \\
				\midrule
				\multirow{2}{*}{GraphWaveNet}
				& Inflow  & 3.300 & 1.984 & 0.842 & 0.470 & 8.412 & -- \\
				& Outflow & 3.154 & 1.938 & 0.855 & 0.458 & 7.595 & -- \\
				\midrule
				\multirow{2}{*}{ZI-ASTGCN (w/o LLM)}
				& Inflow  & 3.201 & 1.396 & 0.852 & 0.428 & 7.854 & 3.653 \\
				& Outflow & 3.200 & 1.385 & 0.851 & 0.424 & 7.755 & 3.470 \\
				\midrule
				\rowcolor[gray]{0.95}
				\multirow{2}{*}{\textbf{ZI-ASTGCN (Ours)}}
				& Inflow  & \textbf{3.112} & \textbf{1.360} &
				\textbf{0.860} & \textbf{0.418} & \textbf{7.631} &
				\textbf{3.359} \\
				\rowcolor[gray]{0.95}
				& Outflow & \textbf{3.118} & \textbf{1.341} &
				\textbf{0.858} & \textbf{0.411} & 7.679 &
				\textbf{3.076} \\
				\bottomrule
			\end{tabular}%
		}
	\end{table}
	
\begin{figure}[htbp]
		\centering
		\includegraphics[width=\columnwidth]{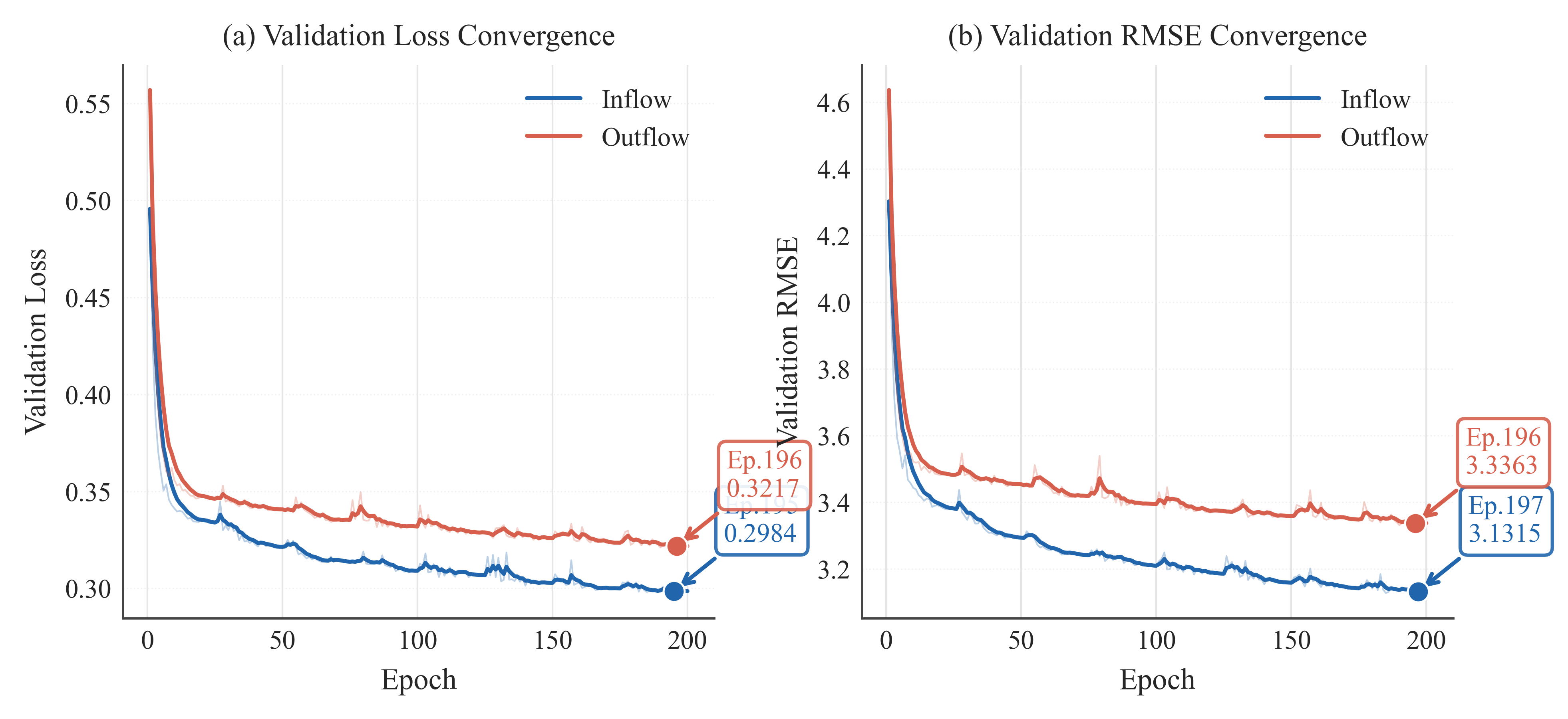}
		\caption{Validation loss (Huber-BCE) and real-scale RMSE over training epochs.}
		\label{fig:loss_curve}
	\end{figure}

	Figure~\ref{fig:loss_curve} traces the training dynamics.  The combined Huber-BCE validation loss decreases monotonically, reaching its minimum at epoch 197 for inflow (0.2984) and epoch 196 for outflow (0.3217); the corresponding real-scale validation RMSE settles at 3.131 and 3.336, respectively.  The alignment between the regression and classification loss profiles throughout training indicates that the multi-task objective does not induce gradient interference between the gating and regression branches, a concern that can undermine dual-head architectures when the two tasks operate on very different signal scales.
	
	\subsection{Event-Driven Demand Anticipation}
	
\begin{figure}[htbp]
		\centering
		\includegraphics[width=\columnwidth]{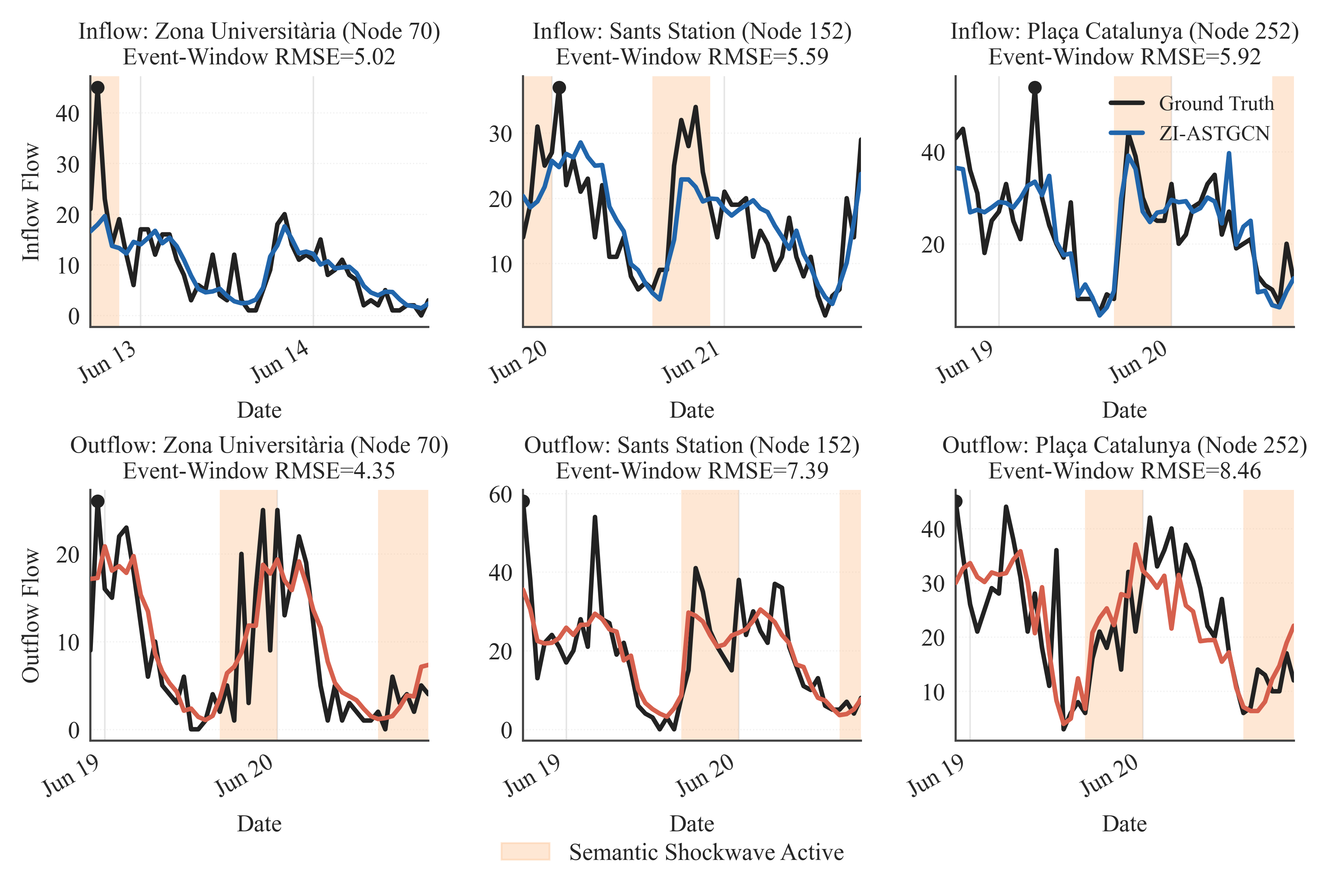}
		\caption{ZI-ASTGCN forecasts against ground-truth demand at three urban nodes during active semantic shockwave intervals (orange shading).}
		\label{fig:timeseries_events}
	\end{figure}

	Figure~\ref{fig:timeseries_events} shows time-series forecasts at three functionally distinct nodes: Zona Universit\`aria (Node 70, a university campus), Sants Station (Node 152, a major intermodal hub), and Pla\c{c}a Catalunya (Node 252, the principal city-center square). The plotted windows are centered on intervals of active semantic shockwaves (shaded in orange), each corresponding to a large scheduled event in the input corpus.
	
	During these windows, observed demand (solid black) rises sharply and departs from the periodic baseline. Autoregressive models are inherently reactive: they begin to adjust only after the anomaly enters the input buffer, accumulating a phase delay proportional to the look-back window length.  ZI-ASTGCN, informed by the shockwave fields ahead of time, tracks both the onset and the peak amplitude of each surge with substantially lower event-period RMSE.  The contrast is most pronounced at Pla\c{c}a Catalunya, where foot traffic spillover from large concerts is not spatially proximate to any single transit terminal, making it poorly handled by topology-only models.
	
	\subsection{Error Distribution Across Traffic Regimes}
	
\begin{figure}[htbp]
		\centering
		\includegraphics[width=\columnwidth]{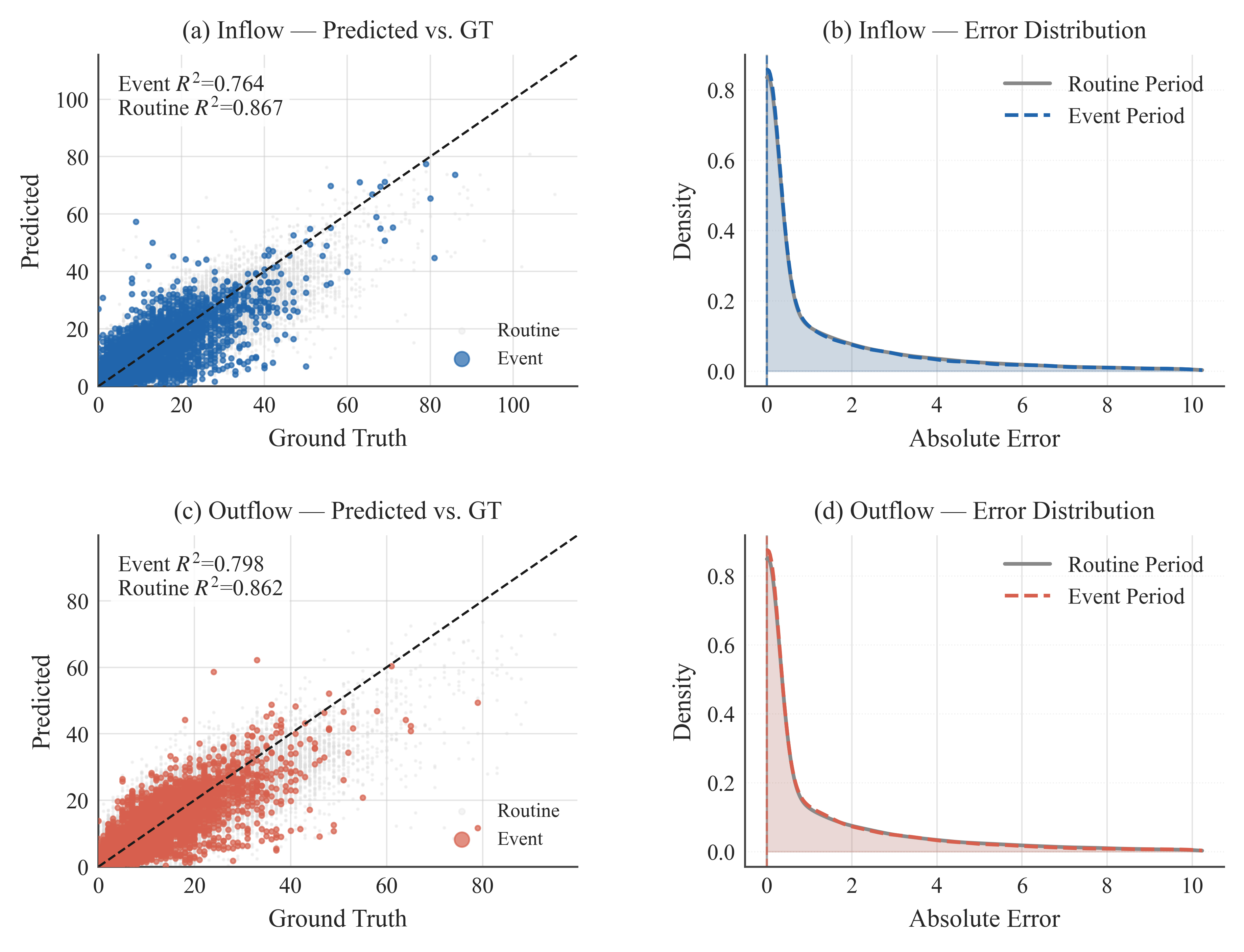}
		\caption{Stratified error analysis across traffic regimes. (a)(c) Predicted vs.\ actual flow during routine and event periods. (b)(d) KDE of absolute residuals for the two regimes.}
		\label{fig:event_analysis}
	\end{figure}

	Figure~\ref{fig:event_analysis} evaluates whether the semantic module induces any distributional shift in prediction errors. The scatter plots in panels (a) and (c) show that ZI-ASTGCN preserves a tight alignment of the point clouds around the identity line for both routine and event intervals. During events, the $R^2$ values are 0.764 (inflow) and 0.798 (outflow), staying within 10 percentage points of the routine period $R^2$ (0.867 and 0.862). This indicates that the added input complexity in event windows does not meaningfully impair the structural goodness of fit.
	
	Panels (b) and (d) present kernel density estimates overlaying the absolute residual distributions for routine versus event periods. In both cases, the distributions are zero-centered and leptokurtic, with nearly identical support and peak density. First, the lack of a rightward displacement in the event-period residual curve shows that the shockwave features do not induce systematic overprediction under normal conditions—a typical failure mode when exogenous inputs encode spurious signals outside their intended domain. Second, the similar tail mass across regimes implies that the tail-sensitive regression branch compresses error dispersion uniformly, rather than preferentially shrinking variance only during event periods.
	
	\subsection{Spatial Error Distribution and Robustness}
	
\begin{figure}[htbp]
		\centering
		\includegraphics[width=\columnwidth]{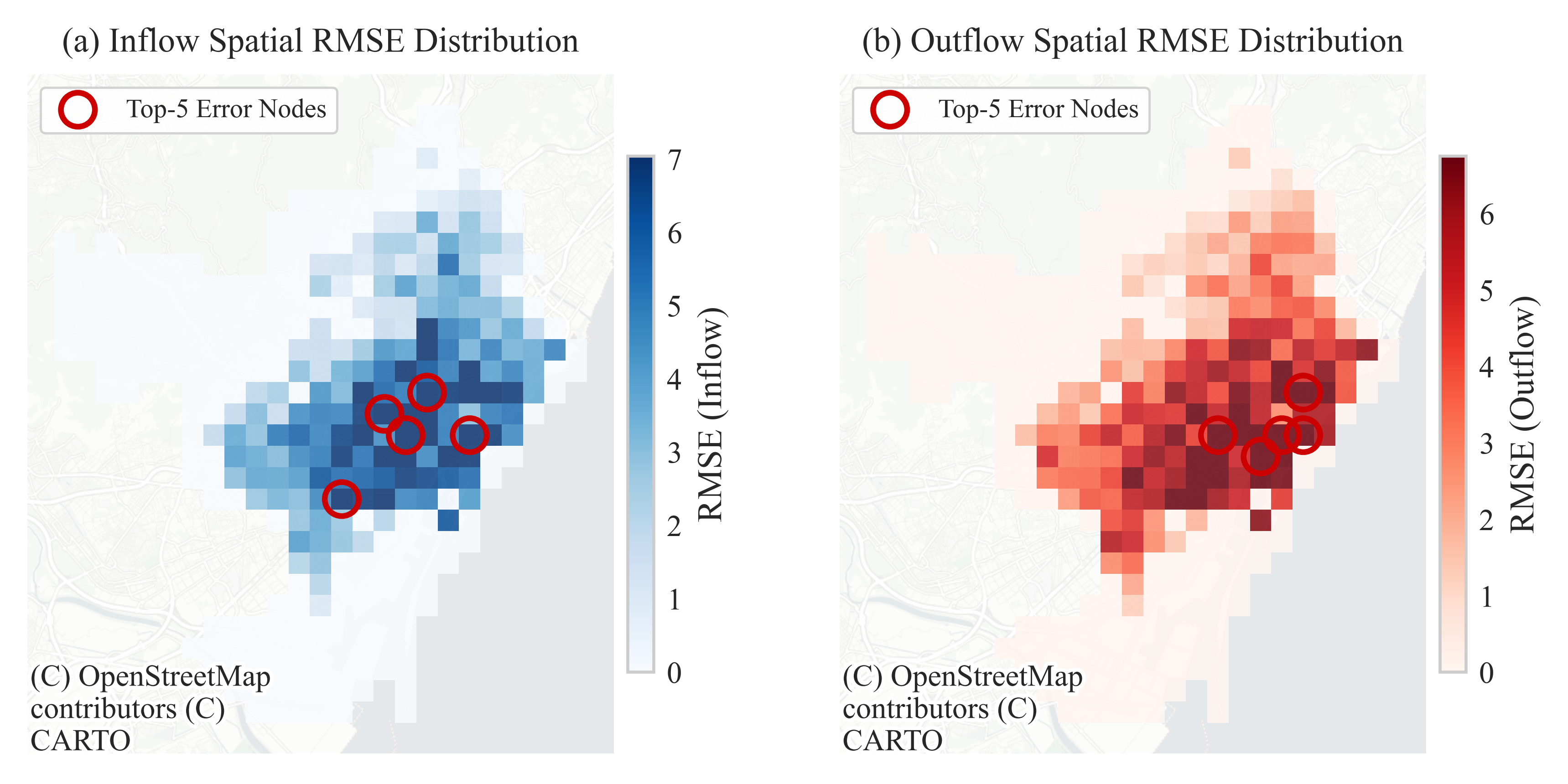}
		\caption{Geographic distribution of grid-level RMSE across Barcelona.}
		\label{fig:spatial_rmse}
	\end{figure}

	Figure~\ref{fig:spatial_rmse} maps grid-level RMSE across Barcelona. Prediction errors peak near Sants Station and the central commercial corridor, where absolute traffic volumes are highest—an expected outcome given that RMSE grows proportionally with flow magnitude. Inflow errors reach at most 7 and outflow errors at most 6 within these hubs, yet both decay sharply in surrounding cells.
	
	The steep spatial decay of errors has a structural interpretation: the dynamic graph fusion mechanism, by weighting adjacency connections on the basis of real-time spatial attention rather than fixed proximity alone, confines non-stationary surge variance to the cell where it originates.  Without this adaptive rewiring, prediction errors would propagate laterally through the graph, inflating RMSE in cells that share a structural edge with a surge node but are not themselves affected by the event.  The spatial maps show no such lateral contamination, validating the design choice to condition $\mathbf{A}_{\mathrm{fused}}$ on both semantic inputs and the current traffic state.
	
	\section{Conclusion and Future Work}
	\label{sec:conclusion}
	
	 This study presented an LLM-augmented spatio-temporal framework for grid-level bike-sharing demand forecasting to address the limited capability of conventional graph-based models in capturing demand disturbances induced by urban events. The proposed semantic shockwave mechanism converts textual event descriptions into parameterized Gaussian decay fields and integrates them into a ZI-ASTGCN through cross-attention and a dual-branch zero-inflated output layer, enabling the model to jointly account for sparse observations and event-driven demand surges within a unified architecture. Experiments on the Barcelona Bicing dataset showed consistent improvements over representative spatio-temporal neural baselines, with larger gains observed during event periods, while ablation analysis indicated that semantic event representations contributed measurable performance improvements beyond the graph structure alone. These results suggest that incorporating structured, physically motivated semantic features can improve event-aware spatio-temporal forecasting. Nevertheless, several limitations remain. The current LLM-based parameter extraction is conducted offline and may limit deployment in real-time applications; additional data sources, such as video streams and social media signals, were not considered and may improve the detection of unscheduled events; and the generalizability of the proposed shockwave representation across cities with different urban forms and bike-sharing systems requires further validation
	
	\section*{Acknowledgments}
	This research has been partially funded by the Spanish Ministry of Science, Innovation and Universities (Ministerio de Ciencia, Innovación y Universidades, Gobierno de España), Grant No. PID2024-159028OB-I00.
	
	\bibliographystyle{IEEEtran}
	\bibliography{root}
	
\end{document}